\documentclass[a4paper,conference]{IEEEtran}
\pdfoutput=1
\usepackage{graphicx}
\usepackage{overpic}
\usepackage{comment}
\usepackage{bm}
\usepackage{amsmath,amssymb} 

\usepackage{dsfont} 
\usepackage{color}
\usepackage[dvipsnames]{xcolor}
\usepackage{multirow}
\usepackage[]{algorithm2e}
\renewcommand{\baselinestretch}{0.97}

\ifCLASSINFOpdf
\else
\fi
\begin{document}
%
\title{Shape Consistent 2D Keypoint Estimation \\ under Domain Shift}



%
\author{\IEEEauthorblockN{Levi O. Vasconcelos\IEEEauthorrefmark{1},
Massimiliano Mancini\IEEEauthorrefmark{2}\IEEEauthorrefmark{3},
Davide Boscaini\IEEEauthorrefmark{4},\\
Samuel Rota Bul\`o\IEEEauthorrefmark{5},
Barbara Caputo\IEEEauthorrefmark{6,7} and
Elisa Ricci\IEEEauthorrefmark{1}\IEEEauthorrefmark{4}}
\IEEEauthorblockA{\IEEEauthorrefmark{1}University of Trento,\IEEEauthorrefmark{2}Sapienza University of Rome,}
\IEEEauthorblockA{\IEEEauthorrefmark{3}University of T\"ubingen,\IEEEauthorrefmark{4}Fondazione Bruno Kessler,\IEEEauthorrefmark{5}Mapillary Research,}
\IEEEauthorblockA{\IEEEauthorrefmark{6}Politecnico di Torino,\IEEEauthorrefmark{7}Italian Institute of Technology}
}


\maketitle

\begin{abstract}
Recent unsupervised domain adaptation methods based on deep architectures have shown remarkable performance not only in traditional classification tasks but also in more complex problems involving structured predictions (e.g. semantic segmentation, depth estimation). Following this trend, in this paper we present a novel deep adaptation framework for estimating keypoints under \textit{domain shift}, i.e. when the training (\textit{source}) and the test (\textit{target}) images significantly differ in terms of visual appearance. Our method seamlessly combines three different components: feature alignment, adversarial training and self-supervision. Specifically, our deep architecture leverages from domain-specific distribution alignment layers to perform target adaptation at the feature level. Furthermore, a novel loss is proposed which combines an adversarial term for ensuring aligned predictions in the output space and a geometric consistency term which guarantees coherent predictions between a target sample and its perturbed version. 
Our extensive experimental evaluation conducted on three publicly available benchmarks shows that our approach outperforms state-of-the-art domain adaptation methods in the 2D keypoint prediction task.
\end{abstract}


%
\IEEEpeerreviewmaketitle

\section{Introduction}
Human behaviour analysis is of utmost importance in many applications such as health care, surveillance, marketing, etc. It is therefore unsurprising that, over the years, researchers have invested significant efforts into developing computational methods able to automatically describe the behavior of people, both at individual and group level. One of the fundamental tasks in computer vision when analysing humans is the problem of accurately estimating 2D/3D keypoints from images depicting face and human bodies \cite{tulsiani2015viewpoints,jakab2018unsupervised,thewlis2019unsupervised}.

Traditional approaches for detecting landmarks from visual inputs operate in a supervised learning setting \cite{andriluka2018posetrack,mpii_andriluka20142d}, thus assuming the availability of large-scale datasets for training with abundant annotations. To sidestep this limitation, 
recently, some methods have proposed deep learning-based models which attempt to predict face and body landmarks in an unsupervised fashion \cite{jakab2018unsupervised,siarohin2019animating,thewlis2019unsupervised}. 

Inspired by these works, in this paper we also address the problem of predicting 2D keypoints locations from static images when no supervision is provided in the domain of interest (\textit{target domain}). However, we assume to have access to annotated data of an auxiliary domain (\textit{source domain}) and cast this problem under an unsupervised domain adaptation (UDA) framework \cite{csurka2017domain}. This assumption is very reasonable as source domain data can be easily derived from pre-existing datasets or from synthetically generated humans obtained with simulators for which annotations are inherently available \cite{zimmermann2017learning}. The main advantage of considering an UDA setting is the possibility to obtain semantically meaningful keypoints (e.g. keypoints associated to specific face and body parts), something that even the more sophisticated unsupervised landmark discovery techniques cannot achieve. Another prominent advantage of casting keypoint estimation under an UDA framework is the possibility to learn an accurate prediction model even when few samples are available in the target domain.

Domain adaptation methods develop from the idea of learning in presence of
\emph{domain shift} or \emph{dataset bias} \cite{torralba2011unbiased}, i.e. when there exists a discrepancy between the training (\emph{source}) and test (\emph{target}) data distributions.
This discrepancy negatively affects the performance of visual recognition systems and the model trained on source data typically is not effective when 
tested in the target samples. In the context of keypoint estimation, the domain shift 
may be observed not only when going from synthetic human models to real data, but also when train and test data correspond to different modalities (RGB vs depth) or when data are collected in different environmental conditions (e.g. different places, different times of the day).

To cope with the domain shift, over the years several UDA approaches \cite{csurka2017domain} have been proposed, both using shallow and deep models. 
Most previous works on UDA focused on object recognition tasks.
Typical strategies for coping with domain shift attempt to align the distributions of the features learned at different levels of the deep architecture by minimizing appropriate moment matching losses \cite{carlucci2017autodial,roy2019unsupervised}, by considering adversarial learning objectives \cite{hong2018conditional,tzeng2017adversarial,ganin2016domain} or by employing deep generative models which synthesize labeled target samples \cite{Bousmalis:Google:CVPR17,hoffman2017cycada,russo2018source}.
While most previous works on UDA mostly addressed categorization problems, recently the computer vision community have started to investigate the possibility of developing adaptation algorithms specialized to structured output prediction problems.
Notable works include those developing UDA methods for semantic segmentation \cite{hong2018conditional,zhang2017curriculum,hoffman2016fcns,chen2017no} or depth prediction \cite{zhang2019online}. 

\begin{figure*}[t]
    \centering
    \includegraphics[width=1\textwidth, trim=80 500 70 45, clip]{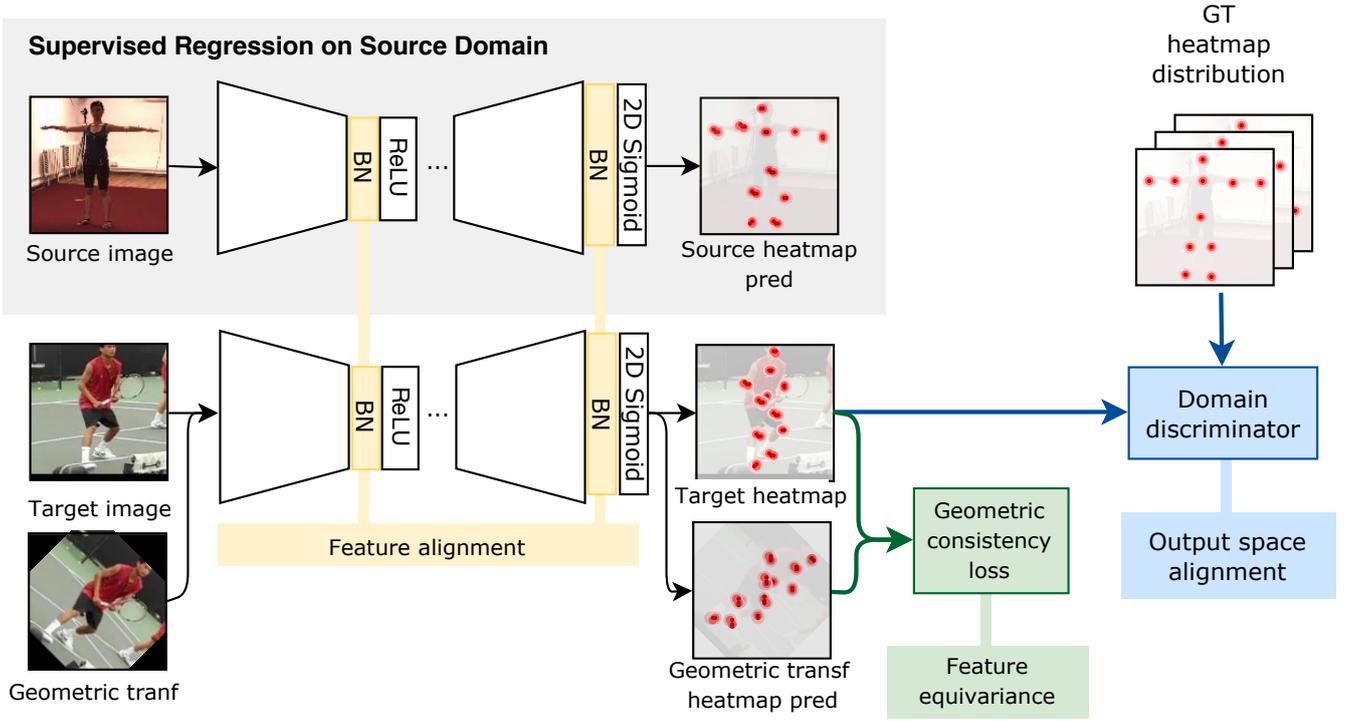}
    \vspace{-5pt}
    \caption{Overview of the proposed unsupervised domain adaptation method. The grey box denotes the source pre-processing step, where the model is pre-trained with annotations from the source domain. In the adaptation step, three complementary components are integrated: feature alignment (yellow), output space alignment (blue) and feature equivariance (green).}
    \label{fig:method}
    \vspace{-0.5cm}
\end{figure*}

In this paper we follow this trend and we develop a novel UDA approach for detecting 2D keypoints from images under domain shift. Our approach operates by considering a pre-trained keypoint detector obtained from source data. Then, when target data are available, adaptation is achieved thanks to three main components. First, features are aligned by introducing an alignment procedure derived from batch normalization \cite{carlucci2017autodial}. Second, a discriminator is employed to estimate the likelihood of the output prediction to be consistent with a given human shape prior (e.g. source ground truth annotations). Thirdly, 
a geometric consistency term is introduced to ensure coherent part-based predictions among corresponding samples under different transformations.
An overview of our method is depicted in Fig.~\ref{fig:method}. 
We evaluate our approach for predicting 2D keypoints on several publicly available benchmarks: Human3.6M \cite{h36m_pami}, Leeds Sports Pose (LSP) Dataset \cite{lsp_johnson2010clustered}, MPII Human Pose dataset \cite{mpii_andriluka20142d} and PennAction \cite{zhang2013actemes}. Experiments show that we outperform previous UDA methods based on adversarial training \cite{tzeng2017adversarial} and domain-alignment layers \cite{carlucci2017autodial}. 

\textbf{Contributions.} To summarize, the main contributions of this work are: 
\begin{itemize}
    \item We introduce a lightweight UDA approach for estimating 2D keypoints from static images which has the favourable property of not requiring source samples during the adaptation phase.
\item We show how a geometric consistency loss can be seamlessly integrated within an adversarial framework for achieving accurate and consistent structured predictions even in absence of annotated target samples.
\item We evaluate our approach on several 2D keypoint estimations benchmarks with different conditions and we show that it is more accurate than previous UDA methods. 
\end{itemize}

\section{Related work}
\label{sec:related}

In the following we review previous approaches on UDA, discussing both categorization based and structured prediction models. Since we propose a deep architecture for unsupervised learning of keypoints, we also review related work on deep self-supervised keypoint discovery.

\textbf{Domain Adaptation.}
Unsupervised domain adaptation methods \cite{csurka2017domain}
leverage the knowledge extracted from labeled data in one or multiple
source domains to learn a classification/regression model for a different but related target domain where no labeled data are provided. A crucial issue in DA is how to compensate with the distribution mismatch between source and target distributions.
As claimed by Ben \textit{et al.} \cite{ben2007analysis}, the feature representation space plays an important role in DA approaches. Therefore, it is not surprising that the common practice for aligning source and target data distributions is to seek for a feature mapping that matches them. Following this idea, the two prominent strategies for coping with the domain shift are alignment trough moment matching \cite{carlucci2017autodial,roy2019unsupervised,carlucci2017just} and adversarial training \cite{long2018conditional,hong2018conditional,tzeng2017adversarial,ganin2016domain}.

Moment matching based methods attempt to align source and target data distributions by considering statistical moments at first and second orders. For instance, approached based on Maximum Mean Discrepancy, \textit{i.e.} the distance between the mean of domain feature distributions, are proposed in \cite{long2015learning,long2016deep,venkateswara2017deep,tzeng2014deep}. Other works \cite{sun2016deep,morerio2017minimal,peng2018synthetic} consider second-order statistics and match them by correlation alignment. Other methods \cite{carlucci2017autodial,li2016revisiting,mancini2018boosting,roy2019unsupervised,mancini2019adagraph,mancini2019inferring} propose to integrate the alignment process within the deep architecture and employ domain-specific alignment layers derived from batch normalization (BN) \cite{ioffe2015batch} or whitening transforms \cite{siarohin2018whitening}. 

Adversarial-based techniques attempt to learn domain-invariant representations considering adversarial losses and networks acting as domain discriminators. For instance, in \cite{ganin2014unsupervised} a gradient reversal layer proposed and used to promote the learning of domain-invariant features.  Similarly, in  \cite{Hoffman:Adda:CVPR17} a domain confusion loss is used to align the source and the target domain feature distributions. Generative adversarial networks (GANs) \cite{goodfellow2014generative} are proposed for DA in CyCADA \cite{russo2018source}, I2I Adapt \cite{murez2018image} and Generate To Adapt (GTA) \cite{sankaranarayanan2018generate}. The idea is to consider GANs to create synthetic source and/or target images.
Our approach is related to previous works in the first and second category, as we also leverage from domain-alignment layers to perform adaptation at feature level while we exploit a discriminator to align the output space. 

While recent research on domain adaptation proposed different strategies to align domain distributions, the vast majority of previous works focused on object classification \cite{csurka2017domain}. Surprisingly, despite their relevance for many computer vision tasks, UDA methods for structured output prediction problems have received little attention so far. Some works have considered the problem of 
semantic segmentation \cite{hong2018conditional,zhang2017curriculum,hoffman2016fcns,chen2017no}, proposing adversarial techniques for coping with the domain shift.
Other works have extended this approach to another pixel-level prediction problem, i.e. depth prediction \cite{zhang2019online}. 

Recently, some works introduced an approach for 3D keypoints estimation \cite{Zhou2018,vasconcelos2019structured}. In particular, in \cite{Zhou2018} the DA problem in the context of 3D keypoints prediction is solved by optimizing a loss made of two terms: a consistency term and a Chamfer distance term. 
The consistency term enforces that the predicted 3D keypoints from different views of the same input must be spatially coherent. The Chamfer loss has the purpose to align the posterior distributions of source and target datasets.
While our approach also employs a geometric consistency term and a term acting on prediction level, the design of our losses is radically different. Finally, differently from \cite{Zhou2018}, source samples are not needed during the adaptation phase, thus enabling knowledge transfer in a lightweight manner.
In \cite{vasconcelos2019structured} an approach which combines domain distribution alignment in the feature space with prior losses on the output space is also presented. In particular, the authors propose to collect statistics about 3D keypoints positions from source training data and to use them as prior information to constrain predictions on adaptation time by introducing a loss derived from Multidimensional Scaling. This works shed some similarity with our method as it also consider adaptation at feature level and in the output space. However, our formulation radically differ from their multidimensional scaling method. Furthermore, it is worth noting that \cite{Zhou2018,vasconcelos2019structured} cannot be applied in our setting since they focus on 3D rather than 2D keypoint detection.





\textbf{Self-supervised 2D Keypoints Estimation.} 
Automatically estimating keypoints from visual data is a problem of utmost importance in computer vision. Keypoints are important in several tasks. For instance, they permit to establish correspondence between different instance of objects of the same category or different viewpoints. Traditional methods for landmark detection consider a supervised setting. Given a specific object category and a large scale dataset of manually annotated keypoints a deep architecture is trained to predict their location by minimizing an L1/L2 loss.

While this approach has led to accurate predictions even in case of difficult datasets \cite{andriluka2018posetrack,mpii_andriluka20142d}, to overcome time-consuming and expensive data annotation, recently the research community have started to develop methods for unsupervised landmark discovery.
For instance, Jakab \textit{et al.} \cite{jakab2018unsupervised} propose a network with an encoder-decoder structure with a differentiable keypoint bottleneck. Zhang \textit{et al.}  \cite{zhang2018unsupervised} introduce an approach to discover  keypoints from a single images given information about frame difference in term of optical flow. Both methods use different loss and regularization terms to compensate for the lack of supervision. Siarohin \textit{et al.} \cite{siarohin2019animating} propose a different self-supervised technique for landmark discovery exploiting video frames and motion streams, showing how their approach can be used for object animation. 

While effective, all these work have a fundamental drawback: keypoints are not semantically meaningful as a consequence of the entire loss of sources of supervision. The lack of semantics on the landmarks prohibits the use of the metrics adopted throughout this work to evaluate both our proposed methods as well as the baselines. For that reason, we do not provide direct comparisons with unsupervised methods.

Recently, \cite{sanchez2019object} consider an adaptation framework for unsupervised landmark discovery. Assuming a pre-trained landmark detector, which has already learned a structured representation from supervised data, they propose to adapt it with unsupervised target data from another domain, with possibly different semantics, by updating a small set of model parameters. Our approach develops from a similar intuition, as we also consider a pre-trained source model and adapt its parameters accordingly to the target distribution. 
However, our purpose is different as well as our approach, since we assume source and target data to share the same semantic and we impose different and more tailored loss constraints exploiting this assumption. For this reason, we compare our approach with DA baselines sharing the same assumptions, showing experimentally the effectiveness of our choices.

\section{Method}
\label{sec:method}
The goal of this work is to adapt a pre-trained 2D keypoint estimation model to a new visual domain for which only unlabelled data are available. Formally, let us denote as $f_s:\mathcal{X}\rightarrow\mathcal{Y}$ our keypoint prediction model, mapping inputs from the image space $\mathcal{X}$ into a set of 2D keypoint coordinates in $\mathcal{Y}\subset \Re^{k\times2}$, where $k$ denotes the number of keypoints. In the following we assume that $f_s$ is a deep neural network that has been pre-trained on a \textit{source} dataset $\mathcal{S}=\{(x^s_1,{y}^s_1),\dots,(x^s_n,{y}^s_n)\}$ where $(x^s_i,y_i^s)$ denotes an image-label pair. Defining $\mathcal{Y}^s=\{y_i^s | (x_i^s,y_i^s)\in\mathcal{S}\}$, we also assume that each $y^s\in\mathcal{Y}^s$ is a set of \textit{ordered} keypoints, with all elements of $\mathcal{Y}^s$ sharing the same order. Moreover, we are given a set of unlabelled data $\mathcal{T}=\{x^t_1,\dots,x^t_{m}\}$ we want to adapt our model to. In the following, we denote $\mathcal{T}$ as our \textit{target} domain. Moreover, we consider $\mathcal{T}$ and $\mathcal{S}$ to have two different joint distributions over $\mathcal{X}\times \mathcal{Y}$ i.e. $p^\mathcal{S}(\mathtt{x},\mathtt{y})\neq p^\mathcal{T}(\mathtt{x},\mathtt{y})$. Given $\mathcal{T}$ and auxiliary prior regarding the output space structure $\mathcal{Y}^s$, our goal is to learn a function $f_t$ starting from the parameters $f_s$ that can effectively predict keypoints from images in $\mathcal{T}$, overcoming the different distributions of the two domains, i.e. the \textit{domain-shift} problem.

In order to solve this problem, we choose to start by revisiting a simple yet effective strategy for DA in image-level classification, based on domain alignment layers \cite{carlucci2017autodial,carlucci2017just}. In particular, this strategy performs a two-level alignment. The first level is the feature one, where the goal is aligning the two marginal distributions $p^\mathcal{S}(\mathtt{x})$ and $p^\mathcal{T}(\mathtt{x})$. The second is the output-level, where priors on the conditional distribution $p^\mathcal{S}(\mathtt{y}|\mathtt{x})$ are used to align $p^\mathcal{T}(\mathtt{y}|\mathtt{x})$ to $p^\mathcal{S}(\mathtt{y}|\mathtt{x})$. In the following, we will first describe \cite{carlucci2017just}, which serves as starting point of our work (Sec.~\ref{sec:method-dial}). We then describe how we revisited the output-level alignment of \cite{carlucci2017just} for the task of 2D keypoint estimation (Sec.~\ref{sec:method-structured-dial}) in a structured fashion, by means of an adversarial objective. We then illustrate how the same semantic objective can be reinforced by using geometric consistency (Sec.~\ref{sec:method-geometric-consistency}). Finally we describe how our approach can be easily applied even when $\mathcal{S}$ is either partially available or replaced by priors about the structure of $p^\mathcal{S}(\mathtt{y})$ (Sec.~\ref{sec:method-continuous}).

\subsection{Feature-level Adaptation with Domain Alignment Layers}
\label{sec:method-dial}
In \cite{carlucci2017just}, the authors proposed a strategy for performing DA in the context of classification by means of domain alignment layers replacing the standard batch-normalization (BN) layers \cite{ioffe2015batch} commonly used within deep architectures. In particular, let us denote as $z$ the features extracted by our model at a given layer, channel and spatial location from an image $x\in\mathcal{D}$ with $\mathcal{D}$ being a domain (in our case, $\mathcal{D}\in\{\mathcal{S},\mathcal{T}\}$). A domain alignment layer (DAL) works as follows:
\begin{equation}
    \label{eq:dal}
    \hat{z}=\text{DAL}^\mathcal{D}(z) = \gamma \frac{z-\mu_\mathcal{D}}{\sqrt{\sigma^2_\mathcal{D} + \epsilon}}+\beta
\end{equation}
where $\mu_\mathcal{D}$ and $\sigma_\mathcal{D}$ are the mean and standard deviation for the given layer, channel and spatial location computed using only samples of $\mathcal{D}$, while $\gamma$ and $\beta$ are scale and bias components. Note that, differently from standard BN, here the normalization statistics $\mu_\mathcal{D}$ and $\sigma_\mathcal{D}$ are \textit{domain-specific}. Normalizing each sample with its {domain-specific} statistics, forces each channel to be aligned to a reference distribution (i.e. the standard normal).
In practice, we can obtain $f_t$ from $f_s$ by computing domain-specific statistics for $\mathcal{T}$ and replacing each BN layer in $f_s$ with $\text{DAL}^\mathcal{T}$, similarly to \cite{li2016revisiting,mancini2018kitting}. 

One missing point from the DAL layers, is the fact that the alignment considers only the marginal distributions $p^\mathcal{S}(\mathtt{x})$ and $p^\mathcal{T}(\mathtt{x})$, while they do not take into account the shift among the conditional distributions $p^\mathcal{S}(\mathtt{y}|\mathtt{x})$ and  $p^\mathcal{T}(\mathtt{y}|\mathtt{x})$. To solve this issue, in \cite{carlucci2017just,carlucci2017autodial}, the predictions given by $f_t$ are used as prior to measure the uncertainty we have on the semantic of the target samples. Under this perspective, we can transfer semantic information from $\mathcal{S}$ to $\mathcal{T}$ by minimizing the uncertainty on the predictions of our model on target samples. In particular, in the case of classification, this means minimizing the classical entropy loss \cite{carlucci2017just,carlucci2017autodial,morerio2017minimal} and consistency-based variants \cite{roy2019unsupervised}. While these strategies are suitable for discrete output spaces, they cannot be directly applied to the 2D keypoint estimation task due to the continuous and structured nature of our output space, which would make them intractable. In the following we will explain how we modeled the uncertainty on the predictions on target samples by means of adversarial training and geometric consistency. 

\subsection{Structured Adversarial 2D Keypoints Alignment}
\label{sec:method-structured-dial}
As highlighted in \cite{morerio2017minimal}, using only the semantic supervision obtained on target data from uncertainty-based losses is not enough for domain adaptation and should be coupled with techniques performing feature-level alignment. This means that, if we want to move the DA strategy of \cite{carlucci2017just,carlucci2017autodial} from image-level classification to 2D keypoint estimation, we need to model the two components together, providing both feature-level and semantic-level alignment. For this reason, while we can still employ Eq.~\eqref{eq:dal} to perform feature-level alignment, we must replace the uncertainty-based losses used in classification with more suitable components that capture the reliability of the global structure of the 2D keypoints predicted by the network while being still computationally tractable.

In order to do this, we adopt an adversarial strategy to discern reliable 2D keypoint configurations from unrealistic ones. In particular, we use the GAN framework \cite{Goodfellow:GAN:NeurIPS2014}, treating our prediction model $f_t$ as the generator and instantiating a new function $h$ acting as our discriminator. Following previous works \cite{jakab2018unsupervised}, apply the discriminator not directly into the output space $\mathcal{Y}$ but into an auxiliary and semantically richer space $\mathcal{H}$, i.e. $h:\mathcal{H}\rightarrow\{0,1\}$. Formally, we will use the Least-Squares GAN formulation of \cite{mao2017least}, defining the discriminator loss as:
\begin{align}
   \label{eq:loss-discrim}
    \mathcal{L}_{disc} =& \mathbb{E}_{y^s\sim\mathcal{Y}^s}[(1-h_r(y^s))^2] \\&+ \mathbb{E}_{x^t\sim\mathcal{T}} [h_r(f_t(x^t))^2]  
\end{align}
and the generator loss as:
\begin{equation}
   \label{eq:loss-gen}
    \mathcal{L}_{gen} = \mathbb{E}_{x^t\sim\mathcal{T}} [(1-h_r(f_t(x^t)))^2]
\end{equation}
with $h_r=h\circ r$ where $r:\mathcal{Y}\rightarrow\mathcal{H}$ is a non-parametric function projecting the set of keypoint coordinates into the auxiliar representation in $\mathcal{H}$. As auxiliar representation we choose the skeleton images of \cite{jakab2018unsupervised}, defining $r$ as the projections of keypoints into skeleton images (see Eq. (1) in \cite{jakab2018unsupervised}). While any representation can be chosen for the keypoints configuration, we resort to the skeleton images \cite{jakab2020self} since they allow to i) explicitly encode relations among keypoints, through the skeleton bones and ii) we can convert the problem into an image-to-image translation one which is easier to solve. Moreover, we can employ a multi-scale discriminator which is crucial to capture the reliability of the predicted keypoint configurations and providing a good supervision signal for target samples, as we will show in our experiments.


\subsection{Geometric Consistency for Local Semantic Coherency}
\label{sec:method-geometric-consistency}
The adversarial loss alone is not enough to ensure that the keypoints found are correlated with low-level image structures. Indeed, the network might learn to output a keypoint configuration which is valid for the discriminator but unrelated to the actual content of the input image. 
In order to strengthen the link among the predicted keypoints and the image, we make use of geometrical consistency through a geometric equivariance loss. Given a geometric transformation $\mathcal{G}: \mathcal{X} \rightarrow \mathcal{X}$ (e.g. rotation), we can enforce the estimated keypoints to be equivariant with respect to $\mathcal{G}$ by ensuring that:
\begin{equation}
    f_\Theta(\mathcal{G}(x)) = \mathcal{G}(f_\Theta(x))
\end{equation}
For that, we optimize our network to also minimize the geometric loss $\mathcal{L}_{geo}$:
\begin{equation}
    \mathcal{L}_{geo} = \sum_{i=1}^m|| f_\Theta(\mathcal{G}(x_i^t)) - \mathcal{G}(f_\Theta(x_i^t))||_1
\end{equation}

By enforcing equivariance, we push the network towards looking for distinctive structures within the input image. This helps the adaptation process when combined with the adversarial loss $\mathcal{L}_{disc}$. Note that the source human poses distribution is not necessarily equal to the true (and unknown) target one: by driving the keypoint predictor to focus on informative low-level regions while also optimizing for global shape consistency, we are able to adapt to both texture (input) and pose (output) distributions simultaneously. Thus, the total optimized loss is then:


\vspace{-0.2cm}
\begin{align}
\label{eq:full}
\begin{split}
\mathcal{L}_{total} = \mathcal{L}_{disc} &+ \mathcal{L}_{gen} + \mathcal{L}_{geo}
\end{split}
\end{align}



\subsection{Adaptation without the Source}
\label{sec:method-continuous}

From Eq.~\eqref{eq:full} it is worth highlighting how the only components which strictly need source data is $\mathcal{L}_{disc}$.
Indeed, in $\mathcal{L}_{disc}$ the source labels are used to provide priors regarding valid keypoint configurations to the discriminator.
This limits the practical applicability of the method, since in the reality we may have not always access to the original set $\mathcal{S}$ as well as its labels $\mathcal{Y}^s$.

With this in mind, we should find a shape prior to replace $\mathcal{Y}^s$ in Eq.~\eqref{eq:loss-discrim} and Eq.~\eqref{eq:loss-gen}. 
However, in our context, it is pretty easy to get access to valid human skeletons or sketched human body representations which can serve as our shape prior. 

Formally, let us suppose we have a set of human representations $\mathcal{Q}$, representing our prior about valid keypoint configurations and that $\mathcal{Q}\subset\hat{\mathcal{H}}$, i.e. all elements $q\in\mathcal{Q}$ belong to a shared auxiliary representation $\hat{\mathcal{H}}$. Under the assumption that there exist a mapping $\hat{r}$ between the representation $\hat{\mathcal{H}}$ and our semantic space $\mathcal{H}$ (e.g. image skeletons), i.e. $\hat{r}:\hat{\mathcal{H}}\rightarrow{\mathcal{H}}$, we can define the discrimination loss as:
\begin{align}
   \label{eq:loss-discrim-no-source}
    \mathcal{L}^q_{disc} =& \mathbb{E}_{q\sim\mathcal{Q}}[(1-h_{\hat{r}}(q))^2] \\ &+ \mathbb{E}_{x^t\sim\mathcal{T}} [h_{r}(f_t(x^t))^2]  
\end{align}
Note that the transformation $\hat{r}$ is applied in the first term while $r$ in the second, since the predictions of our network and our prior $\mathcal{Q}$ might belong to different semantic spaces. 

By replacing $\mathcal{L}^{disc}$ with $\mathcal{L}^q_{disc}$ in Eq.~\eqref{eq:full}, we can obtain a new formulation, where there is no dependence on the source prior $\mathcal{Y}^s$.  
Indeed, we want to highlight that, with the latter formulation, no components explicitly requires the presence of the original source set $\mathcal{S}$ used to pre-train the model, making the whole adaptation procedure lightweight and easy to employ in practical scenarios,  even without the presence of source domain $\mathcal{S}$.

\section{Experimental Results}
In this section we provide the results of our experimental evaluation. We first show the results of an ablation study to evaluate the impact on the performance of our technical contributions. Then, we quantitatively and qualitatively compare our model with state of the art adaptation methods.




\subsection{Experimental setup}

\textbf{Datasets.} We evaluate our method on the 
following 
datasets: Human3.6M \cite{h36m_pami}, Leeds Sports Pose (LSP) Dataset \cite{lsp_johnson2010clustered}, MPII Human Pose dataset \cite{mpii_andriluka20142d}, and PennAction \cite{zhang2013actemes}. We use the Humans3.6m dataset as our source domain and the other datasets as target domains.

The Human3.6M \cite{h36m_pami} dataset contains $3.6$M of images of eleven different actors performing 
different action sequences. 
The ground truth was computed using motion capture devices. In our experiments we consider the protocol two described in \cite{sun2017compositional} and use the seven labelled subjects. In particular, independently on the specific setting, we use subjects (S1, S5, S6) as training set of the source domain, following previous works \cite{vasconcelos2019structured}. 
For the output space we consider all 32 joints. 

LSP \cite{lsp_johnson2010clustered} is composed by 2,000 RGB images of full body poses. The dataset is divided into two subsets: training and testing. Each subject has 1,000 images respectively. In our experiments we the training set during the adaptation phase, reporting the performances on the test split. 

MPII Human Pose dataset \cite{mpii_andriluka20142d} is a large dataset with 40,522 images collected from youtube videos. In our experiments we use a 28,821 images for training and 11,701 for testing as suggested by \cite{mpii_andriluka20142d}. As for LSP, in our experiments we use only the training set as target domain during the adaptation phase, reporting the results on the test set.

The PennAction dataset \cite{zhang2013actemes} is a dataset composed of 2326 in-the-wild videos of human performing 15 challenging actions such as golf swing and bowling. The annotations are done frame-by-frame in each video sequences: 13 human body joints were manually annotated with 2D image coordinates. In our work, we use PennAction’s training and testing split for evaluation, which consists of an even split of the videos between training and testing. Since our model works for single images, we do not use any time or sequence information, keeping single frames as input for the model. Again, we use only the images of training set as target domain during training, reporting the results on the test split.

\begin{figure*}[t!]
    \centering
    \scalebox{0.5}{
    \hspace{-4pt}
    \begin{overpic}[width=0.3\textwidth]{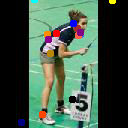}
    \end{overpic}
    }
    \put(-55, 80){\small source}
    \put(118, 80){\small source}
    \put(280, 80){\small source}
    \put(30,80){\small ours}
    \put(206,80){\small ours}
    \put(365,80){\small ours}
    \scalebox{0.5}{
    \hspace{4pt}
    \begin{overpic}[width=0.3\textwidth]{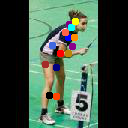}
    \end{overpic}
     }
    \scalebox{0.5}{ 
    \begin{overpic}[width=0.3\textwidth]{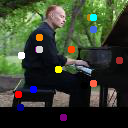}
    \end{overpic}
    }
    \scalebox{0.5}{ 
    \begin{overpic}[width=0.3\textwidth]{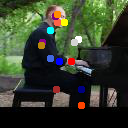}
    \end{overpic}
    }
    \scalebox{0.5}{
    \begin{overpic}[width=0.3\textwidth]{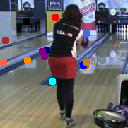}
    \end{overpic}
    }
    \scalebox{0.5}{
    \begin{overpic}[width=0.3\textwidth]{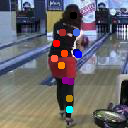}
    \end{overpic}
    }    
    \scalebox{0.5}{
    \begin{overpic}[width=0.3\textwidth]{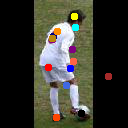}
    \end{overpic}
    }
    \scalebox{0.5}{
    \begin{overpic}[width=0.3\textwidth]{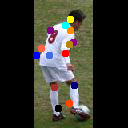}
    \end{overpic}
     }
    \scalebox{0.5}{ 
    \begin{overpic}[width=0.3\textwidth]{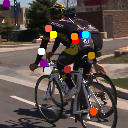}
    \end{overpic}
    }
    \scalebox{0.5}{ 
    \begin{overpic}[width=0.3\textwidth]{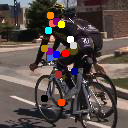}
    \end{overpic}
    }
    \scalebox{0.5}{
    \begin{overpic}[width=0.3\textwidth]{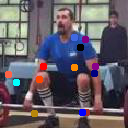}
    \end{overpic}
    }
    \scalebox{0.5}{
    \begin{overpic}[width=0.3\textwidth]{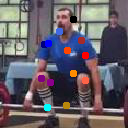}
    \end{overpic}
    }    
    \scalebox{0.5}{
    \begin{overpic}[width=0.3\textwidth]{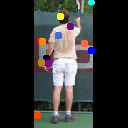}
    \end{overpic}
    }
    \scalebox{0.5}{
    \begin{overpic}[width=0.3\textwidth]{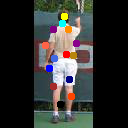}
    \end{overpic}
     }
    \scalebox{0.5}{ 
    \begin{overpic}[width=0.3\textwidth]{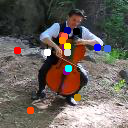}
    \end{overpic}
    }
    \scalebox{0.5}{ 
    \begin{overpic}[width=0.3\textwidth]{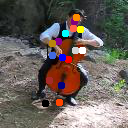}
    \end{overpic}
    }
    \scalebox{0.5}{
    \begin{overpic}[width=0.3\textwidth]{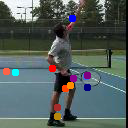}
    \end{overpic}
    }
    \scalebox{0.5}{
    \begin{overpic}[width=0.3\textwidth]{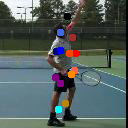}
    \end{overpic}
    }
    \scalebox{0.5}{
    \begin{overpic}[width=0.3\textwidth]{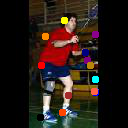}
    \end{overpic}
    }
    \scalebox{0.5}{
    \begin{overpic}[width=0.3\textwidth]{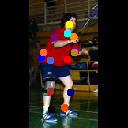}
    \end{overpic}
     }
    \scalebox{0.5}{ 
    \begin{overpic}[width=0.3\textwidth]{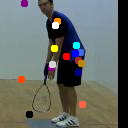}
    \end{overpic}
    }
    \scalebox{0.5}{ 
    \begin{overpic}[width=0.3\textwidth]{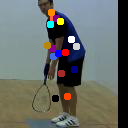}
    \end{overpic}
    }
    \scalebox{0.5}{
    \begin{overpic}[width=0.3\textwidth]{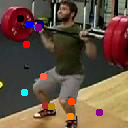}
    \end{overpic}
    }
    \scalebox{0.5}{
    \begin{overpic}[width=0.3\textwidth]{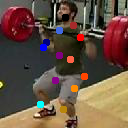}
    \end{overpic}
    }
    \put(-425, -12){LSP}
    \put(-263, -12){MPII}
    \put(-93, -12){PENN}
	\caption{Qualitative comparison of our method and the source model in three different datasets.}
	\label{fig:qualitative-datasets}
\end{figure*}

\textbf{Data Pre-processing.}
For all datasets the samples were pre-processed by cropping, centering and resizing to $128 \times 128$ resolution. We cropped the images using a loose bounding box centered on the foreground human.

\textbf{Training protocols.}
Our keypoint detector model follows an hourglass architecture \cite{newell2016stacked}, where it downsamples and upsamples the input multiple times using residual blocks \cite{he2016deep} to produce heatmaps. We train the keypoint detector initially by doing supervised heatmap regression using the Humans3.6m dataset as source domain. We use the spatial resolution of the heatmaps equal to the input resolution $128 \times 128$.
Our experiments were performed using a GeForce RTX 2080 Ti, the code was implemented in PyTorch framework \cite{paszke2017automatic} and the results reported are averaged computed on 5 runs. 

Since we know the correspondence between our predicted keypoints and their semantic meaning, we chose to evaluate our method similarly to supervised techniques. Note that this is not possible with self-supervised keypoint detection methods \cite{jakab2018unsupervised,zhang2019online} since they do not preserve keypoint semantics. For each dataset 
we select on the source domain only the keypoints that semantically corresponds to the ones annotated in the ground-truth of the target dataset. We evaluate our results using two metrics: the Percentage of Correct Keypoints (PCK) and Mean Squared Error (MSE).

\textbf{Implementation Details.} 
In the following we denote with: (1) $C^s_m$ a convolution-ReLU layer with $m$ filters and stride $s$, (2) $CN^s_{m}$ the same as $C^s_m$ with batch normalization before ReLU. (3) $CI_{m}$ the same as $C^{s}_m$ with instance normalization before ReLU. (4) $A_k$ Average pooling layer with kernel $k \times k$, and (5) $I_f$ an bilinear interpolation upsampling by a factor of $f$. Our model is composed of the following components:
\begin{itemize}
    \item Encoder: $CN^1_{64}-A_2-CN^1_{128}-A_2-CN^1_{256}-A_2-CN^1_{512}-A_2-CN^1_{1024}$
    \item Decoder: $I_2-CN^1_{512}-I_2-CN^1_{256}-I_2-CN^1_{128}-I_2-CN^1_{64}-I_2-CN^1_{32}$
    \item Keypoint Detector: follows an hourglass forwarding strategy, architecture is: Encoder-Decoder-$CN^1_{kp}$, where $kp$ stands for the number of keypoints.
    \item Discriminator: We use a multi-scale discriminator, where the following architecture is applied to scales $[1, \frac{1}{2}, \frac{1}{4}]$ of the input image: $CI^2_{64}-CI^2_{128}-CI^2_{256}-CI^2_{512}-CI^2_{1}$
\end{itemize}

We use Adam optimizer (learning rate: $2 \times 10^{-4}$,  $\beta_1 = 0.5$, $\beta_2 = 0.999$), batch size of $16$. The same set of hyperparameters is shared among all experiments and datasets. As the transformation $\mathcal{G}$ for the geometric consistency loss we employ 2D rotation with random angles.


\subsection{Results} 

We first perform a thorough ablation study that highlights the impact of each module separately. We then compare our method with state-of-the-art domain adaptation approaches.


\begin{figure}[t]
    \centering
	\begin{overpic}[width=0.9\columnwidth, trim=0 0 30 35, clip]{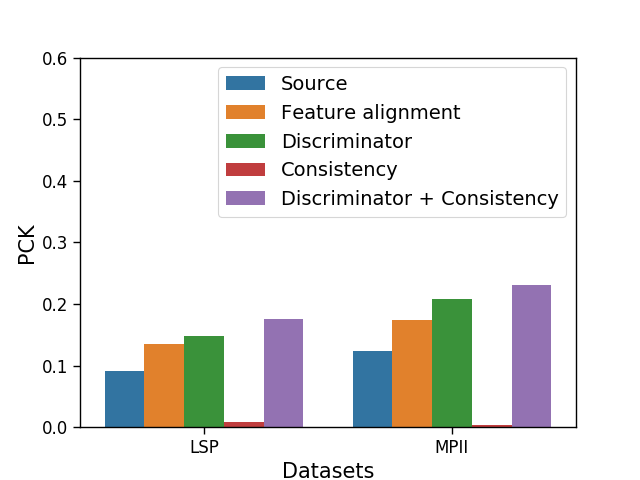} 
    \end{overpic}
    \caption{Ablation study on LSP and MPII datasets. On both datasets the combination of the discriminator and consistency terms shows a significant improvement on adaptation performance.}
    \vspace{-10pt}
    \label{fig:ablation}
\end{figure}

\textbf{Analysis of our approach. }
We start analyzing the performance of our method module-wise, isolating the impact of each design choice for adaptation. The ablation study is performed on two datasets: LSP and MPII. As stated above, on both settings, we consider Humans3.6m as our source domain, and the other datasets as target. 
We compare the following models: (i) the pre-trained source model, (ii) a model performing just the feature alignment step through BN, as in \cite{mancini2018kitting,li2016revisiting}, (iii) a model trained to align the output distributions using a discriminator, (iv) a model enforcing only geometric consistency and (v) our proposed approach, combining all previous modules. 

The results are reported in Figure \ref{fig:ablation}.
As shown in the figure, only applying the BN-based adaptation strategy allows to improve the performances with respect to the baseline model. The discriminator module enforces the output of the model to be of similar distribution as the source ground-truth. By enforcing a prior knowledge about the keypoint shape, we are able to improve further the performance of the model on target domain. We attribute this improvement to the fact that the overall keypoint shape does not change among target and source domains. The geometric consistency loss alone though is not able to converge, since a trivial solution is found by the module where all keypoints are located at the center of the image. Nonetheless, when we combine geometric consistency with the output discriminator module, the model is able to correlate keypoints semantic to meaningful image structures, finding keypoints with higher accuracy.

\begin{table}[t!]
\caption{Ablation study on different discriminator models.}
\centering
\begin{tabular}{lcccc}
\hline
                                  & \multicolumn{2}{c}{LSP}                                & \multicolumn{2}{c}{MPII}                               \\ \hline
Discriminator Type                & \textbf{PCK}    & \textbf{MSE}                         & \textbf{PCK}    & \textbf{MSE}                         \\ \hline
\multicolumn{1}{l|}{Single Scale} & 0.1564          & \multicolumn{1}{c|}{0.1639}          & \textbf{0.2341} & \multicolumn{1}{c}{0.1551}          \\
\multicolumn{1}{l|}{Multi Scale}  & \textbf{0.1764} & \multicolumn{1}{c|}{\textbf{0.1539}} & 0.2337          & \multicolumn{1}{c}{\textbf{0.1547}}\\
\hline
\end{tabular}
\label{tab:disc}
\end{table}

We also performed an ablation on different types of discriminator, namely the impact of using single- vs. multi-scale models. The ablation summarized at Table~\ref{tab:disc} suggests that multi-scale is a more robust model, bringing clear improvements over the single scale one on both datasets and both metrics.

\textbf{Comparison with state of the art.}
In this section we compare our model with state of the art DA methods on three datasets. Among previous domain adaptation methods we choose to compare with AutoDIAL \cite{carlucci2017autodial}, a previous method based on domain alignment layers, and with ADDA  \cite{tzeng2017adversarial}, a state of the art approach considering adversarial training. We choose these methods as they are closely related to our approach. In particular AutoDIAL \cite{carlucci2017autodial} operate by matching distributions at feature level using first and second order statistics. Differently, ADDA  \cite{tzeng2017adversarial} uses a discriminator for aligning domain distributions, similarly to our proposal. Both methods have been implemented taking the original codes from the authors webpage.

The results of our evaluation are provided in Table \ref{tab:comparison}. From the table it it evident that our approach outperforms all previous UDA methods. We ascribe the superior performance of our approach to the combination of multiple adaptations strategies within the same framework.
We show qualitative results of our adaptation method on Figure~\ref{fig:qualitative-datasets}, where we compare our adapted model to the source model. Although all three datasets have different keypoint semantics, we were able to successfully adapt to all three different domains using the same source model. The source model was trained on all 32 keypoints of Humans3.6m dataset, then, only for sake of comparison, for each target dataset we use 
only the keypoints that semantically match the annotations of each target dataset. Note that despite the output space distribution is not exactly the same (e.g. the scale and range of poses of the datasets are different than our prior distribution) our method is flexible enough to adapt and also to keep the semantics for each keypoint inherited from the source model. We can see that the model achieves a good balance between global and local keypoint estimation: the geometric loss drives the model to respond to meaningful low-level image structures, while the adversarial loss empowers the model with global-shape perception. It is interesting to note how, thanks to the geometric constraint, our model is robust to occlusions (e.g. left of the tennis player, first row LSP) and able to tailor the predictions to the actual content of the image even if they violate the shape priors used in the adversarial loss (e.g. sport actions where not included in the source dataset). 

\begin{table}[t!]
\centering
\caption{Comparison with state of the art methods on three datasets}
\scalebox{0.9}{
\begin{tabular}{lcccccc}
\hline
    & \multicolumn{2}{c}{LSP} & \multicolumn{2}{c}{MPII} & \multicolumn{2}{c}{PENN} \\
    \hline
    & \textbf{PCK} & \textbf{MSE} & \textbf{PCK} & \textbf{MSE} & \textbf{PCK} & \textbf{MSE} \\
    \hline
    \multicolumn{1}{l|}{Source Only} & 0.0906 & \multicolumn{1}{c|}{0.3185} & 0.1229 & \multicolumn{1}{c|}{0.2501} & 0.2332 & 0.3882 \\
    \multicolumn{1}{l|}{AutoDIAL \cite{carlucci2017autodial} \ \ \ }                           & 0.1456          & \multicolumn{1}{c|}{0.3240}          & 0.1845          & \multicolumn{1}{c|}{0.2140}          & 0.2731     & 0.3262     \\
    \multicolumn{1}{l|}{ADDA \cite{tzeng2017adversarial}}                           & 0.1497          & \multicolumn{1}{c|}{0.1627}          & 0.2132          & \multicolumn{1}{c|}{0.1734}          & 0.2943     & \textbf{0.2247}     \\
    \multicolumn{1}{l|}{Ours} & \textbf{0.1764} & \multicolumn{1}{c|}{\textbf{0.1539}} & \textbf{0.2337} & \multicolumn{1}{c|}{\textbf{0.1547}} & \textbf{0.3032}     & 0.2264  \\ 
    \hline
\end{tabular}
}
\vspace{-10pt}
\label{tab:comparison}
\end{table}

\section{Conclusions}
In this work we present an approach for 2D human pose estimation under an Unsupervised Domain Adaptation scenario. Our method addresses the domain shift problem by aligning features with batch normalization-based techniques. Moreover, we exploit prior shape information and geometric equivariance to align also the output space, where we constrain our output space to be consistent with our prior knowledge, while also flexible enough to encompass necessary deformations to cope with the target distribution.
We presented quantitative and qualitative experiments in which we show our method effectiveness with respect to state of the art models.






%
\bibliographystyle{IEEEtran}
\bibliography{IEEEfull}

\begin{thebibliography}{10}
\providecommand{\url}[1]{#1}
\csname url@samestyle\endcsname
\providecommand{\newblock}{\relax}
\providecommand{\bibinfo}[2]{#2}
\providecommand{\BIBentrySTDinterwordspacing}{\spaceskip=0pt\relax}
\providecommand{\BIBentryALTinterwordstretchfactor}{4}
\providecommand{\BIBentryALTinterwordspacing}{\spaceskip=\fontdimen2\font plus
\BIBentryALTinterwordstretchfactor\fontdimen3\font minus
  \fontdimen4\font\relax}
\providecommand{\BIBforeignlanguage}[2]{{%
\expandafter\ifx\csname l@#1\endcsname\relax
\typeout{** WARNING: IEEEtran.bst: No hyphenation pattern has been}%
\typeout{** loaded for the language `#1'. Using the pattern for}%
\typeout{** the default language instead.}%
\else
\language=\csname l@#1\endcsname
\fi
#2}}
\providecommand{\BIBdecl}{\relax}
\BIBdecl

\bibitem{tulsiani2015viewpoints}
S.~Tulsiani and J.~Malik, ``{Viewpoints and Keypoints},'' in \emph{Proc. CVPR},
  2015.

\bibitem{jakab2018unsupervised}
T.~Jakab, A.~Gupta, H.~Bilen, and A.~Vedaldi, ``Unsupervised learning of object
  landmarks through conditional image generation,'' in \emph{Advances in Neural
  Information Processing Systems}, 2018, pp. 4016--4027.

\bibitem{thewlis2019unsupervised}
J.~Thewlis, S.~Albanie, H.~Bilen, and A.~Vedaldi, ``Unsupervised learning of
  landmarks by descriptor vector exchange,'' in \emph{Proceedings of the IEEE
  International Conference on Computer Vision}, 2019, pp. 6361--6371.

\bibitem{andriluka2018posetrack}
M.~Andriluka, U.~Iqbal, E.~Insafutdinov, L.~Pishchulin, A.~Milan, J.~Gall, and
  B.~Schiele, ``Posetrack: A benchmark for human pose estimation and
  tracking,'' in \emph{Proceedings of the IEEE Conference on Computer Vision
  and Pattern Recognition}, 2018, pp. 5167--5176.

\bibitem{mpii_andriluka20142d}
M.~Andriluka, L.~Pishchulin, P.~Gehler, and B.~Schiele, ``2d human pose
  estimation: New benchmark and state of the art analysis,'' in
  \emph{Proceedings of the IEEE Conference on computer Vision and Pattern
  Recognition}, 2014, pp. 3686--3693.

\bibitem{siarohin2019animating}
A.~Siarohin, S.~Lathuili{\`e}re, S.~Tulyakov, E.~Ricci, and N.~Sebe,
  ``Animating arbitrary objects via deep motion transfer,'' in
  \emph{Proceedings of the IEEE Conference on Computer Vision and Pattern
  Recognition}, 2019, pp. 2377--2386.

\bibitem{csurka2017domain}
G.~Csurka, \emph{Domain adaptation in computer vision applications}.\hskip 1em
  plus 0.5em minus 0.4em\relax Springer, 2017, vol.~2.

\bibitem{zimmermann2017learning}
C.~Zimmermann and T.~Brox, ``{Learning to Estimate 3D Hand Pose from Single RGB
  Images},'' in \emph{Proc. ICCV}, 2017.

\bibitem{torralba2011unbiased}
A.~Torralba and A.~A. Efros, ``{Unbiased Look at Dataset Bias},'' in
  \emph{Proc. CVPR}, 2011.

\bibitem{carlucci2017autodial}
F.~M. Carlucci, L.~Porzi, B.~Caputo, E.~Ricci, and S.~Rota~Bul{\`o},
  ``{AutoDIAL: Automatic Domain Alignment Layers},'' in \emph{Proc. ICCV},
  2017.

\bibitem{roy2019unsupervised}
S.~Roy, A.~Siarohin, E.~Sangineto, S.~Rota~Bul{\`o}, N.~Sebe, and E.~Ricci,
  ``{Unsupervised Domain Adaptation using Feature-Whitening and Consensus
  Loss},'' in \emph{Proc. CVPR}, 2019.

\bibitem{hong2018conditional}
W.~Hong, Z.~Wang, M.~Yang, and J.~Yuan, ``{Conditional Generative Adversarial
  Network for Structured Domain Adaptation},'' in \emph{Proc. CVPR}, 2018.

\bibitem{tzeng2017adversarial}
E.~Tzeng, J.~Hoffman, K.~Saenko, and T.~Darrell, ``{Adversarial Discriminative
  Domain Adaptation},'' in \emph{Proc. CVPR}, 2017.

\bibitem{ganin2016domain}
Y.~Ganin, E.~Ustinova, H.~Ajakan, P.~Germain, H.~Larochelle, F.~Laviolette,
  M.~Marchand, and V.~Lempitsky, ``{Domain-Adversarial Training of Neural
  Networks},'' \emph{Trans. JMLR}, vol.~17, no.~59, pp. 1--35, 2016.

\bibitem{Bousmalis:Google:CVPR17}
K.~Bousmalis, N.~Silberman, D.~Dohan, D.~Erhan, and D.~Krishnan,
  ``{Unsupervised Pixel-level Domain Adaptation with GANs},'' in \emph{Proc.
  CVPR}, 2017.

\bibitem{hoffman2017cycada}
J.~Hoffman, E.~Tzeng, T.~Park, J.-Y. Zhu, P.~Isola, K.~Saenko, A.~A. Efros, and
  T.~Darrell, ``{CyCADA: Cycle-Consistent Adversarial Domain Adaptation},'' in
  \emph{Proc. ICML}, 2018.

\bibitem{russo2018source}
P.~Russo, F.~M. Carlucci, T.~Tommasi, and B.~Caputo, ``From source to target
  and back: symmetric bi-directional adaptive gan,'' in \emph{CVPR}, 2018.

\bibitem{zhang2017curriculum}
Y.~Zhang, P.~David, and B.~Gong, ``{Curriculum Domain Adaptation for Semantic
  Segmentation of Urban Scenes},'' in \emph{Proc. ICCV}, 2017.

\bibitem{hoffman2016fcns}
J.~Hoffman, D.~Wang, F.~Yu, and T.~Darrell, ``{FCNs in the Wild: Pixel-level
  Adversarial and Constraint-based Adaptation},'' \emph{arXiv:1612.02649},
  2016.

\bibitem{chen2017no}
Y.-H. Chen, W.-Y. Chen, Y.-T. Chen, B.-C. Tsai, Y.-C. Frank~Wang, and M.~Sun,
  ``{No More Discrimination: Cross City Adaptation of Road Scene Segmenters},''
  in \emph{Proc. ICCV}, 2017.

\bibitem{zhang2019online}
Z.~Zhang, S.~Lathuili{\`e}re, A.~Pilzer, N.~Sebe, E.~Ricci, and J.~Yang,
  ``{Online Adaptation through Meta-Learning for Stereo Depth Estimation},''
  \emph{arXiv:1904.08462}, 2019.

\bibitem{h36m_pami}
C.~Ionescu, D.~Papava, V.~Olaru, and C.~Sminchisescu, ``{Human3.6M: Large Scale
  Datasets and Predictive Methods for 3D Human Sensing in Natural
  Environments},'' \emph{IEEE Trans. PAMI}, vol.~36, no.~7, pp. 1325--1339,
  2014.

\bibitem{lsp_johnson2010clustered}
S.~Johnson and M.~Everingham, ``Clustered pose and nonlinear appearance models
  for human pose estimation.'' in \emph{bmvc}, vol.~2, no.~4.\hskip 1em plus
  0.5em minus 0.4em\relax Citeseer, 2010, p.~5.

\bibitem{zhang2013actemes}
W.~Zhang, M.~Zhu, and K.~G. Derpanis, ``From actemes to action: A
  strongly-supervised representation for detailed action understanding,'' in
  \emph{Proceedings of the IEEE International Conference on Computer Vision},
  2013, pp. 2248--2255.

\bibitem{ben2007analysis}
S.~Ben-David, J.~Blitzer, K.~Crammer, and F.~Pereira, ``{Analysis of
  Representations for Domain Adaptation},'' in \emph{Proc. NeurIPS}, 2007.

\bibitem{carlucci2017just}
F.~M. Carlucci, L.~Porzi, B.~Caputo, E.~Ricci, and S.~Rota~Bul{\`o}, ``{Just
  DIAL: DomaIn Alignment Layers for Unsupervised Domain Adaptation},'' in
  \emph{Proc. ICIAP}, 2017.

\bibitem{long2018conditional}
M.~Long, Z.~Cao, J.~Wang, and M.~I. Jordan, ``{Conditional Adversarial Domain
  Adaptation},'' in \emph{Proc. NeurIPS}, 2018.

\bibitem{long2015learning}
M.~Long and J.~Wang, ``{Learning Transferable Features with Deep Adaptation
  Networks},'' in \emph{Proc. ICML}, 2015.

\bibitem{long2016deep}
M.~Long, H.~Zhu, J.~Wang, and M.~I. Jordan, ``Deep transfer learning with joint
  adaptation networks,'' \emph{ICML}, 2017.

\bibitem{venkateswara2017deep}
H.~Venkateswara, J.~Eusebio, S.~Chakraborty, and S.~Panchanathan, ``Deep
  hashing network for unsupervised domain adaptation,'' in \emph{CVPR}, 2017.

\bibitem{tzeng2014deep}
E.~Tzeng, J.~Hoffman, N.~Zhang, K.~Saenko, and T.~Darrell, ``Deep domain
  confusion: Maximizing for domain invariance,'' \emph{arXiv preprint
  arXiv:1412.3474}, 2014.

\bibitem{sun2016deep}
B.~Sun and K.~Saenko, ``Deep coral: Correlation alignment for deep domain
  adaptation,'' \emph{arXiv preprint arXiv:1607.01719}, 2016.

\bibitem{morerio2017minimal}
P.~Morerio, J.~Cavazza, and V.~Murino, ``{Minimal-Entropy Correlation Alignment
  for Unsupervised Deep Domain Adaptation},'' in \emph{Proc. ICLR}, 2018.

\bibitem{peng2018synthetic}
X.~Peng and K.~Saenko, ``Synthetic to real adaptation with generative
  correlation alignment networks,'' in \emph{WACV}, 2018.

\bibitem{li2016revisiting}
Y.~Li, N.~Wang, J.~Shi, J.~Liu, and X.~Hou, ``Revisiting batch normalization
  for practical domain adaptation,'' \emph{arXiv preprint arXiv:1603.04779},
  2016.

\bibitem{mancini2018boosting}
M.~Mancini, L.~Porzi, S.~Rota~Bul{\`o}, B.~Caputo, and E.~Ricci, ``{Boosting
  Domain Adaptation by Discovering Latent Domains},'' in \emph{Proc. CVPR},
  2018.

\bibitem{mancini2019adagraph}
M.~Mancini, S.~R. Bul{\`o}, B.~Caputo, and E.~Ricci, ``Adagraph: Unifying
  predictive and continuous domain adaptation through graphs,'' \emph{CVPR},
  2019.

\bibitem{mancini2019inferring}
M.~Mancini, L.~Porzi, S.~R. Bulo, B.~Caputo, and E.~Ricci, ``Inferring latent
  domains for unsupervised deep domain adaptation,'' \emph{IEEE T-PAMI}, 2019.

\bibitem{ioffe2015batch}
S.~Ioffe and C.~Szegedy, ``{Batch Normalization: Accelerating Deep Network
  Training by Reducing Internal Covariate Shift},'' in \emph{Proc. ICML}, 2015.

\bibitem{siarohin2018whitening}
A.~Siarohin, E.~Sangineto, and N.~Sebe, ``Whitening and coloring batch
  transform for gans,'' \emph{ICLR}, 2018.

\bibitem{ganin2014unsupervised}
Y.~Ganin and V.~Lempitsky, ``{Unsupervised Domain Adaptation by
  Backpropagation},'' \emph{Proc. ICML}, 2015.

\bibitem{Hoffman:Adda:CVPR17}
E.~Tzeng, J.~Hoffman, T.~Darrell, and K.~Saenko, ``Adversarial discriminative
  domain adaptation,'' in \emph{CVPR}, 2017.

\bibitem{goodfellow2014generative}
I.~Goodfellow, J.~Pouget-Abadie, M.~Mirza, B.~Xu, D.~Warde-Farley, S.~Ozair,
  A.~Courville, and Y.~Bengio, ``Generative adversarial nets,'' in \emph{NIPS},
  2014.

\bibitem{murez2018image}
Z.~Murez, S.~Kolouri, D.~Kriegman, R.~Ramamoorthi, and K.~Kim, ``Image to image
  translation for domain adaptation,'' in \emph{CVPR}, 2018.

\bibitem{sankaranarayanan2018generate}
S.~Sankaranarayanan, Y.~Balaji, C.~D. Castillo, and R.~Chellappa, ``Generate to
  adapt: Aligning domains using generative adversarial networks,'' in
  \emph{CVPR}, 2018.

\bibitem{Zhou2018}
X.~Zhou, A.~Karpur, C.~Gan, L.~Luo, and Q.~Huang, ``{Unsupervised Domain
  Adaptation for 3D Keypoint Estimation via View Consistency},'' in \emph{Proc.
  ECCV}, 2018.

\bibitem{vasconcelos2019structured}
L.~O. Vasconcelos, M.~Mancini, D.~Boscaini, B.~Caputo, and E.~Ricci,
  ``Structured domain adaptation for 3d keypoint estimation,'' in \emph{2019
  International Conference on 3D Vision (3DV)}.\hskip 1em plus 0.5em minus
  0.4em\relax IEEE, 2019, pp. 57--66.

\bibitem{zhang2018unsupervised}
Y.~Zhang, Y.~Guo, Y.~Jin, Y.~Luo, Z.~He, and H.~Lee, ``Unsupervised discovery
  of object landmarks as structural representations,'' in \emph{Proceedings of
  the IEEE Conference on Computer Vision and Pattern Recognition}, 2018, pp.
  2694--2703.

\bibitem{sanchez2019object}
E.~Sanchez and G.~Tzimiropoulos, ``Object landmark discovery through
  unsupervised adaptation,'' in \emph{Advances in Neural Information Processing
  Systems}, 2019, pp. 13\,498--13\,509.

\bibitem{mancini2018kitting}
M.~Mancini, H.~Karaoguz, E.~Ricci, P.~Jensfelt, and B.~Caputo, ``{Kitting in
  the Wild through Online Domain Adaptation},'' \emph{IROS}, 2018.

\bibitem{Goodfellow:GAN:NeurIPS2014}
I.~Goodfellow, J.~Pouget-Abadie, M.~Mirza, B.~Xu, D.~Warde-Farley, S.~Ozair,
  A.~Courville, and Y.~Bengio, ``Generative adversarial nets,'' in
  \emph{NeurIPS}, 2014.

\bibitem{mao2017least}
X.~Mao, Q.~Li, H.~Xie, R.~Y. Lau, Z.~Wang, and S.~Paul~Smolley, ``Least squares
  generative adversarial networks,'' in \emph{Proceedings of the IEEE
  International Conference on Computer Vision}, 2017, pp. 2794--2802.

\bibitem{jakab2020self}
T.~Jakab, A.~Gupta, H.~Bilen, and A.~Vedaldi, ``Self-supervised learning of
  interpretable keypoints from unlabelled videos,'' in \emph{CVPR}, 2020.

\bibitem{sun2017compositional}
X.~Sun, J.~Shang, S.~Liang, and Y.~Wei, ``{Compositional Human Pose
  Regression},'' in \emph{Proc. ICCV}, 2017.

\bibitem{newell2016stacked}
A.~Newell, K.~Yang, and J.~Deng, ``Stacked hourglass networks for human pose
  estimation,'' in \emph{European conference on computer vision}.\hskip 1em
  plus 0.5em minus 0.4em\relax Springer, 2016, pp. 483--499.

\bibitem{he2016deep}
K.~He, X.~Zhang, S.~Ren, and J.~Sun, ``{Deep Residual Learning for Image
  Recognition},'' in \emph{Proc. CVPR}, 2016.

\bibitem{paszke2017automatic}
A.~Paszke, S.~Gross, S.~Chintala, G.~Chanan, E.~Yang, Z.~DeVito, Z.~Lin,
  A.~Desmaison, L.~Antiga, and A.~Lerer, ``{Automatic Differentiation in
  PyTorch},'' in \emph{Proc. NeurIPS-WS}, 2017.

\end{thebibliography}

\end{document}